\documentclass[letterpaper, 10 pt, conference]{ieeeconf} 

\IEEEoverridecommandlockouts                              
\usepackage{graphicx}
\usepackage{amsmath} 
\usepackage{amssymb}  

\usepackage{algorithm}
\usepackage[noend]{algpseudocode}

\usepackage [english]{babel}
\usepackage [autostyle, english = american]{csquotes}
\MakeOuterQuote{"}

\makeatletter
\let\NAT@parse\undefined
\makeatother
\usepackage[hidelinks]{hyperref}

\makeatletter
\def\BState{\State\hskip-\ALG@thistlm}
\makeatother
\title{\LARGE \bf
A Hitchhiker's Guide On Distributed Training of Deep Neural Networks
}

\author{Karanbir Chahal$^{1}$, Manraj Singh Grover$^{2}$ and Kuntal Dey$^{3}$
\thanks{$^{1}$ Karanbir is an independent researcher }
\thanks{$^{2}$ Manraj is an independent researcher }
\thanks{$^{3}$ Kuntal Dey is an IBM Researcher }}

\begin{document}

\maketitle
\thispagestyle{empty}
\pagestyle{empty}


\begin{abstract}
Deep learning has led to tremendous advancements in the field of Artificial Intelligence. One caveat however is the substantial amount of compute needed to train these deep learning models. Training a benchmark dataset like ImageNet on a single machine with a modern GPU can take upto a week, distributing training on multiple machines has been observed to drastically bring this time down. Recent work has brought down ImageNet training time to a time as low as 4 minutes by using a cluster of 2048 GPUs. This paper surveys the various algorithms and techniques used to distribute training and presents the current state of the art for a modern distributed training framework. More specifically, we explore the synchronous and asynchronous variants of distributed Stochastic Gradient Descent, various All Reduce gradient aggregation strategies and best practices for obtaining higher throughout and lower latency over a cluster such as mixed precision training, large batch training and gradient compression.
\end{abstract}

\section{Introduction}
\label{sec:intro}

\subsection{Background and Motivation}

Data is being generated at an unprecedented scale. Internet scale companies generate terabytes of data every day which needs to be analyzed effectively to draw meaningful insights \cite{fbdata}. Deep Learning has emerged as a powerful tool for performing this analysis, these algorithms boast state of the art results on complex tasks for vision \cite{krizhevsky2012imagenet}, language \cite{devlin2018bert} and intelligent reasoning \cite{mnih2013playing}. Unfortunately, these algorithms need large amounts of data for effective training which takes a substantial amount of time. The first deep learning algorithms that got state of the art results on the ImageNet classification task took a week to train on a single GPU. This speed is no longer sustainable in today's day and age where models need to be trained on data which dwarfs the ImageNet dataset in size. There is an intrinsic need to scale deep learning training in a horizontal manner while also retaining the accuracy of a single GPU model. The speed of this training should ideally decrease linearly with the increase in the number of machines while also being fault tolerant and able to converge under high latency network conditions.

\subsection{Distributed Training Overview}

Distributing training of neural networks can be approached in two ways- data parallelism and model parallelism. Data parallelism seeks to divide the dataset equally onto the nodes of the system where each node has a copy of the neural network along with it's local weights. Each node operates on a unique subset of the dataset and updates it's local set of weights. These local weights are shared across the cluster to compute a new global set of weights through an accumulation algorithm. These global weights are distributed to all the nodes from whereon the processing of the next batch of data commences. 

Model parallelism on the other hand seeks to distribute training by splitting the architecture of the model onto separate nodes. AlexNet \cite{krizhevsky2012imagenet} was one of the first models which used model parallelism by dividing the network among 2 GPU's to fit the model into memory. Model Parallelism is applicable when the model architecture is too big to fit on a single machine and the model has some parts that can be parallelized. Model parallelization is used with some models such as Object Detection Networks \cite{redmon2017yolo9000} which have separate bounding and class prediction heads that are independent of each other. Generally, most networks can fit on 2 GPU's which limits the amount of scalability that can be achieved, hence we primarily focus on data parallelism in this paper.

The paper is roughly divided into six sections, the first section surveys the existing optimization training algorithms and the second section focuses on communication strategies used to connect nodes across the network. The third section explores techniques like large batch size training, gradient compression and mixed precision training for training efficiently with low powered devices and slow network conditions. The fourth section assimilates the information from the previous sections and selects the optimal training algorithm and communication primitive for different settings. The final two sections are divided into the future work and the conclusion.

\section{Components of a Distributed Training Framework}

\subsection{Distributed Training Algorithms}

A popular algorithm used for training in the distributed setting is the Stochastic Gradient Descent (SGD), this algorithm shall be the focal point in our discussion going forward. It is important to note that the principles mentioned for SGD can be easily ported to other popular optimization algorithms such as Adam \cite{kingma2014adam}, RMSProp \cite{tieleman2012lecture} among others \cite{ruder2016overview}.
Distributed SGD algorithms can be roughly classified into two variants- Asynchronous and Synchronous SGD. Synchronous SGD \cite{goyal2017accurate} aims to replicate the algorithm as is in a distributed setting thereby tightly coupling the nodes in the network. On the other hand, Asynchronous SGD \cite{dean2012large} decouples the nodes from other worker nodes by decreasing their interdependence. Although this decoupling allows for greater parallelization, it has an unfortunate side effect of being slightly inferior in stability and accuracy. Several modifications to Asynchronous SGD have been proposed to close this accuracy gap with Synchronous SGD.
Recent trends have gravitated towards scaling Synchronous SGD, more specifically, training networks with large batch sizes has led to promising results. Large mini-batch sizes have a few benefits, the chief one being that SGD over large mini batches allow the model to take bigger steps towards the local minima thereby speeding up the optimization procedure. However in practice, training networks with large batch sizes leads to divergence problems or a "generalization gap" i.e the test accuracy of the network is at times lower than on a model trained on a lower batch size. Recent efforts have been made to train over large batches by modulating the learning rate proportional to the batch size. It has been empirically found that increasing the batch size is equivalent to decreasing the learning rate \cite{smith2017don} making training with large batch sizes a viable method with the added benefit of lesser total parameter updates to train. Linear learning rate scaling has enabled ImageNet \cite{deng2009imagenet} being trained in an hour \cite{goyal2017accurate} by scaling up the batch size to 8,096. A technique called LARS \cite{ginsburg2018large} allows for the use of batches up to 32k and more recently with a combination of mixed precision training in \cite{jia2018highly}, the ImageNet database was successfully trained in 4 minutes using a batch size of 64k. 

\subsection{Communicating Between Nodes}
There is another important component to Distributed Training which is the communication of data between nodes. This is a mature research topic thanks to the work of GFS \cite{ghemawat2003google}, Hadoop \cite{shvachko2010hadoop} and a number of other distributed file systems. Efficient and bandwidth aware communication between nodes in a peer to peer setting require collective communication primitives \cite{thakur2005optimization} which were first introduced in High performance computing (HPC) systems and brought to the world of deep learning by \cite{gibiansky2017bringing}. Modern deep learning frameworks like TensorFlow \cite{abadi2016tensorflow} and PyTorch use these primitives for the All Reduce procedure as it allows for a efficient transfer of gradients between connected nodes in optimal time. All reduce \cite{thakur2005optimization} has several variants like the Ring All Reduce, Recursive Halfing/Doubling and Binary Blocks algorithm that are used in practice. In distributed training, the computation vs communication has to be kept optimal for efficient horizontal scaling. Training remains optimal if the communication step is efficient and synchronized with the computation of various machines i.e computation should finish at approximately the same time across the cluster. In slow network conditions communication between nodes proves to be the bottleneck. Gradient compression and mixed precision training are promising techniques that can increase overall throughput of the network. Recent work \cite{smith2017cyclical} has discovered that using cyclic learning rates can lead to a 10x reduction in the number of epochs needed to achieve network convergence, making it a promising research avenue in distributed training.

\section*{SGD Variants}
Stochastic Gradient Descent \cite{bottou2010large} is an optimization algorithm used to train neural networks. It is a variation of gradient descent, in that it provides an algorithm to tweaks the weights towards a lower minima after each backpropogation step. SGD differs from vanilla gradient descent as it operates on mini batches instead of individual training examples. It is given as follows:
\begin{equation}
    w_{t+1} = w_{t} - \eta \frac{1}{n}\sum_{x\in B} \nabla l(x, w_t)
\end{equation}

where $w_{t+1}$ are the weights computed for the current batch, $n$ is the number of training examples in the mini batch and $\nabla l(x, w_t)$ are the gradients computed for the previous training example.

For a distributed setting, SGD is classified into roughly two types- the asynchronous variant and the synchronous variant. These two and their variants are explored in detail in the next section.

\section{Synchronous SGD}
\label{sec:sync}
Synchronous SGD is a distributed gradient descent algorithm, it is currently one of the most popular optimizers used to distribute training. Nodes in the network compute gradients on their local batch of data after which each node sends their gradients to a master server. The master accumulates these gradients by averaging them to form the new global set of gradients for the weight update step. These global gradients update the local weights of each node by using the same formula as the single machine SGD after which nodes can start processing the next batch of data. This whole procedure is analogous to computing a forward pass and backpropogation step through a single mini batch of data on a single machine, therefore, Synchronous SGD guarantees convergence. However, there are a few limitations of Synchronous SGD:

    \subsection{Stragglers} In a distributed system, machines can take a long time to return a response. Slow network conditions, failed network requests, machine crashes or even byzantine errors are all possible failures that are common in a distributed network. In this unreliable network, Synchronous SGD due to its tightly coupled nature can take a long time to converge. The machines which take a long time to respond are known as stragglers, \cite{chen2016revisiting} observes that 80\% of the second last gradients arrive in under 2 seconds whereas only 30\% of the final gradients do. Furthermore, the time to collect the final few gradients grows exponentially resulting in wasted idle resources and time expended in waiting for the slowest gradients to arrive. A possible solution to this could be to decrease the number of machines. However, reducing the mini batch size increases the total number of iterations required for convergence, \cite{chen2016revisiting} observes that there is nearly a linear increase in number of iterations required for convergence as the mini batch size is decreased. A popular approach to this problem is to introduce backup replicas that perform the same processing as the worker nodes. The gradient aggregation completes when the gradients are received for the first N machines. The use of backup replicas seeks to lower the probability of machine response delay. According to \cite{jin2016scale}, there is a trade-off between the number of replicas and the time for convergence. It is observed for a 100 machine cluster, the optimal configuration is to have 96 workers and 4 backup replicas.

    \subsection{Synchronization Barrier} Another issue with Synchronous SGD is the synchronization barrier. The synchronization barrier is the amount of time spent in waiting for all the workers to send their gradients before aggregating them. In practice, this can take a long time depending on the machine state and network conditions, training is only as fast as the slowest stragglers. This synchronization barrier can be mitigated to some extent by introducing replicas and using better communication primitives that help alleviate it by utilizing network bandwidth more effectively. However, these solutions don't completely resolve the problem of the barrier due to the nature of how Synchronous SGD is modelled. Asynchronous SGD removes this synchronization barrier, however it brings along a new set of problems that need to be dealt with.

    \subsection{Single Point of Failure} Worker nodes communicate with the master node for all gradient exchanges in the master slave setup of vanilla Synchronous SGD leading to a single point of failure. This single point of failure also lends itself to bandwidth problems as a high number of machines try to communicate with a common machine at the same time. Dean et al \cite{dean2012large} try to address this by introducing parameter servers which act as the masters for a subset of worker nodes but a tree like hierarchy still lends itself to single point failures. Peer to peer communication mechanisms like the All Reduce algorithm remove this single point of failure, they also have an added benefit of providing better utilization of network bandwidth than the master slave edition.

\subsection{Fault Tolerance}
A fault tolerance approach in training with Synchronous SGD has not been addressed in literature as of now to the best of our knowledge. Fault tolerance in deep learning training frameworks is managed by systems like Docker, Kubernetes and Spark that use some form of state management and failure recovery, however that it is an externalized solution. Currently, the vanilla All Reduce algorithm needs to be restarted if a single machine fails. We propose a modification to the All Reduce algorithm inspired by the Raft algorithm that allows it to operate in an unstable environment.

\section{Asynchronous SGD}
\label{sec:async}

Asynchronous SGD is a distributed gradient descent algorithm that allows training multiple model replicas in parallel on different nodes with different subsets of the data. Each model replica requests parameter servers for the global weights, processes a mini batch to calculate gradients and sends them back to the parameter server which updates the global weights accordingly. Since each node computes gradients independently and does not require interaction among each other, they can work at their own pace and have greater robustness to machine failure i.e. if one node fails, other nodes can continue processing thus eliminating the problem of synchronization barrier introduced by Synchronous SGD.

\subsection{Stale Gradients}
However, Asynchronous SGD suffers from the problem of "delayed" or "stale" gradients. Since there is no synchronization between workers while updating global model parameters, some workers could be computing gradients using model weights that may be several gradient steps behind the current version of global weights making convergence slow and not guaranteed. Several studies have analyzed and shared approaches to mitigate this problem. R. Zhang et al.'s \cite{zhang2014asynchronous} proposed algorithm combines merits of Delayed Proximal Gradient algorithm \cite{li2013distributed} and Stochastic Variance Reduced Gradient \cite{johnson2013accelerating} which guarantees convergence to optimal solution at fast linear rate while using constant learning rate. W. Zhang et al's \cite{zhang2015staleness} work on $n$-softsync protocol suggested that instead of waiting for all learners $\lambda$ to complete calculating gradients as in case of Synchronous SGD, we update weights after collecting gradients from atleast $c = \lfloor \frac{\lambda}{n} \rfloor$ learners where $n$ is the hyperparameter controlling staleness of the system and modulating learning rate. Experiments show a convergence rate similar to Synchronous SGD while achieving near linear speed ups on image recognition tasks. Another promising work by S. Zheng et al. \cite{zheng2016asynchronous} proposed Delay Compensated ASGD (DC-ASGD) using Taylor expansion of the gradient function and approximation of Hessian matrix to theoretically prove convergence for convex and non-convex optimization problems. Experiments on image recognition tasks show a good balance between speed and accuracy.



\subsection{Elastic Averaging SGD}

Recently, S. Zhang et al. shared Elastic Averaging SGD (EASGD) algorithm in \cite{zhang2015deep} with asynchronous and momentum-based variants. EASGD allows workers to maintain their own local weights and coordinates work using an elastic force linking a center variable with the computed weights. A quadratic penalty is added to the optimization to ensure that the local workers don't fall away from the center variable given by the equation below:

\begin{equation}
F(x_1, \dots, x_p, \Tilde{x}) = \sum_{i=1}^{p}[f(x_i, X_i) + \frac{\rho}{2}\lvert\lvert x^i_t - \Tilde{x}_t \rvert\rvert^2]
\end{equation}

The amount of exploration is controlled by the term $\rho$ introduced by this penalty. The update rules for parameter $x^i$ and center variable $\Tilde{x}$ are given as follows:

\begin{equation}
x^i_{t+1} = x^i_t - \eta(g^i_t(x^i_t) + \rho(x^i_t - \Tilde{x}_t))
\end{equation}

\begin{equation}
\Tilde{x}_{t+1} = \Tilde{x}_t + \eta\sum_{i=1}^{p}\rho(x^i_t - \Tilde{x}_t)
\end{equation}

Here $x^i_t$ and  $\Tilde{x}_t$ denote the values of $x^i$ and $\Tilde{x}$ at iteration $t$, $g^i_t(x^i_t)$ denotes the gradient with respect to $x^i$ at iteration $t$, $\eta$ is the learning rate and $p$ is the number of workers. Updates are performed by the master server after every $\tau$ units of communication period. The algorithm was tested on image classification tasks and found to have a faster convergence rate compared to the Downpour algorithm proposed in \cite{dean2012large} for all values of $\tau$, it also has a comparatively lower validation error for high values of $\tau$ on the \textit{CIFAR} \cite{krizhevsky2009learning} dataset. It was also observed that the validation error decreased as number of workers increased.

\subsection{Gossiping SGD}

Gossiping SGD \cite{jin2016scale}, considered as a decentralized version of Elastic Averaging SGD, makes use of the average of local weights instead of a center variable, thus eliminating the need of a master server. This work analyzes how asynchronous and synchronous SGD algorithms converge at the beginning and end of training. It was observed that Asynchronous SGD converges faster compared to All-Reduce SGD for large step size and for a smaller step size, Gossiping SGD converges faster than the Elastic Averaging SGD. Of all algorithms, \textit{Synchronous All-Reduce SGD} converges most consistently. For small clusters of size upto 32 nodes, both Elastic Averaging and Gossiping converge faster whereas for large scale training with 100 or more nodes, All-Reduce SGD consistently converges to a higher accuracy solution in comparison. 

In summary, the limitations of Asynchronous SGD are --- a single point of failure due to the master slave setup, greater instability during training due to stale gradients and a lower degree of convergence compared to it's synchronous alternatives.

\section*{Gradient Accumulation}
\label{sec:grad_acc}

Gradient Accumulation algorithms represent an important component of distributed training systems. These algorithms are responsible for accumulating the local gradients from each worker node and distributing the updated global gradients back to the worker nodes. The All Reduce algorithm makes for a very good fit for this functionality and also removes the need for a master server by espousing a peer to peer paradigm for data exchange.

If there are $n$ number of machines and each machine has some data with it, the All Reduce algorithm performs an operation on the aggregation of data from each machine and deposits the resultant value to all the machines in the network. This functionality is a good fit for the Synchronous SGD procedure as it averages all the local gradients and deposits the updated gradients to all the machines. Introducing the ring All Reduce algorithm to deep learning \cite{gibiansky2017bringing} has led distributed training using some form of the All Reduce algorithm for gradient distribution. 

There are quite a few variants of the All Reduce algorithm which have been described in the coming sections. This includes our proposed All Reduce algorithm coined Tolerant All-Reduce which is capable of providing fault tolerance in an unstable networking environment.

\section{Ring Algorithm}
\label{sec:ring}

\begin{figure*}[htb]
\centering
\includegraphics[width=0.5\textwidth]{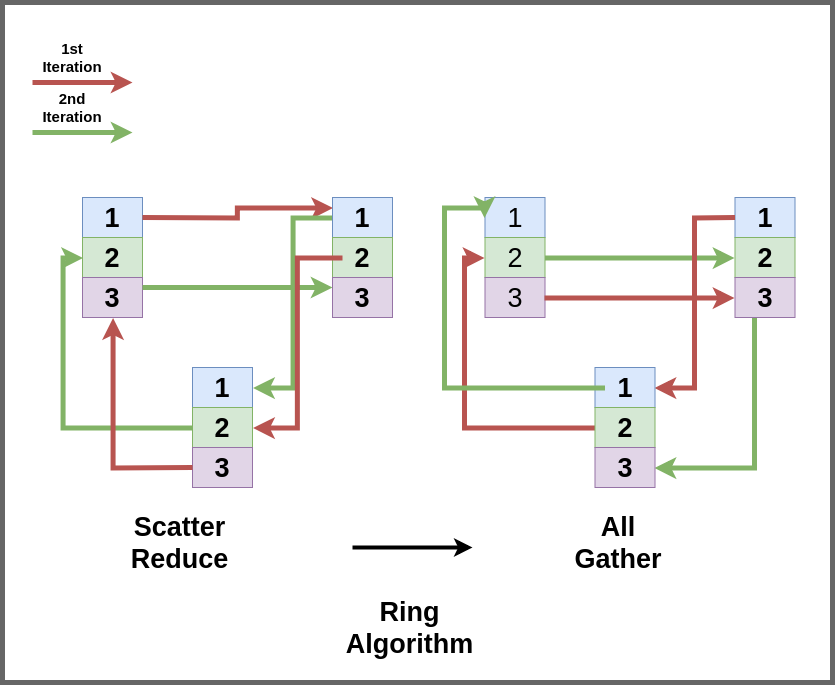}
\label{fig:model2}
\caption{Ring Algorithm}
\end{figure*} 

The ring All Reduce in \cite{thakur2005optimization} works by combining two algorithms- the scatter-reduce and the all gather. 
The scatter reduce algorithm works for $p-1$ steps where $p$ is the number of machines. The gradient vectors are divided into $p$ chunks. The steps of this algorithm are as follows:
\begin{itemize}
\item Each machine at $j^{th}$ step sends the $(i-j+1)^{th}$ chunk to machine $i+1$ and receives $(i-j-1)^{th}$ chunk from machine $i-1$.
\item When a machine receives a value, it performs a reduction procedure with the existing value and stores the new value alongside the original value. This reduced value is sent further and the original value is used as the second operator when a reduction needs to be done.
\item This process carries on until $p-1$ steps. 
\end{itemize}

\begin{algorithm}
\caption{Ring Algorithm}
\begin{algorithmic}[1]
\Procedure{Scatter Reduce}{}
\State $\textit{origValue} \gets \text{gradients computed by machine  }\textit{i}$
\State $redValue \gets \textit{origValue}$
\BState \emph{\textbf{for} j = 0; j \textless p; j++}:
\State $sendToMachine(i -j +1,redValue)$
\State $receivedValue \gets machine(i-j-1)$
\State $redValue \gets receivedValue + origValue$
\EndProcedure
\end{algorithmic}
\end{algorithm}

After the scatter reduce process ends, each machine has a chunk of the final result. Now, each machine simply has to broadcast their piece of the final chunk to all other machines. This is done using the all gather procedure which is very similar to the scatter gather, only instead of a reduction on receiving data, the piece is simply stored as is as it is the final result.

The all gather process is repeated for $p-1$ steps and is given as follows:
\begin{itemize}
\item Each machine at $j^{th}$ step sends $(i-j+1)^{th}$ chunk to process i+1 and receives $(i-j-1)^{th}$ chunk from process $i-1$.
\item When machine receives a value, it stores the value at it's corresponding index.
\item This process carries on with each machine sending it's stored value until $p-1$ steps.
\end{itemize}

\begin{algorithm}
\caption{Ring Algorithm}
\begin{algorithmic}[1]
\Procedure{All Gather}{}
\BState \emph{\textbf{for} j = 0; j \textless p; j++}:
\State $sendToMachine(i -j +1,redValue)$
\State $receivedValue \gets machine(i-j-1)$
\State $redValue \gets receivedValue$
\EndProcedure
\end{algorithmic}
\end{algorithm}

The network latency of this ring All Reduce algorithm is $2*(p-1)$. This algorithm is quite popular and is in use in production grade frameworks like TensorFlow \cite{abadi2016tensorflow} and Caffe. The all ring All Reduce poses the following advantages:
\begin{itemize}
\item \textbf{Efficient Use of Network Bandwidth}: Machines are continuously sending some chunk of data from their machine to another, hence no machines are left idle.
\item \textbf{Peer to peer}: A peer to peer approach ensures that there is no single point of failure.
\item This algorithm is independent of the number of machines i.e it doesn't change it's properties when the number of machines are odd, even, power of twos etc. 
\end{itemize}

However, there are some shortcomings one needs to be aware of when using the ring algorithm. These are given as follows:
\begin{itemize}
\item \textbf{Time Complexity}: The process takes $\mathcal{O}(n)$ time, the algorithms we will study later have $\mathcal{O}(\log{}n)$ complexity.
\item \textbf{Fault Tolerance}: The algorithm is not fault tolerant, if a single machine fails. the whole procedure needs to be started again.
\end{itemize} 

\section{Recursive Halfing and Doubling Algorithm }
\label{sec:halfing}

\begin{figure*}[htb]
\centering
\includegraphics[width=0.7\textwidth]{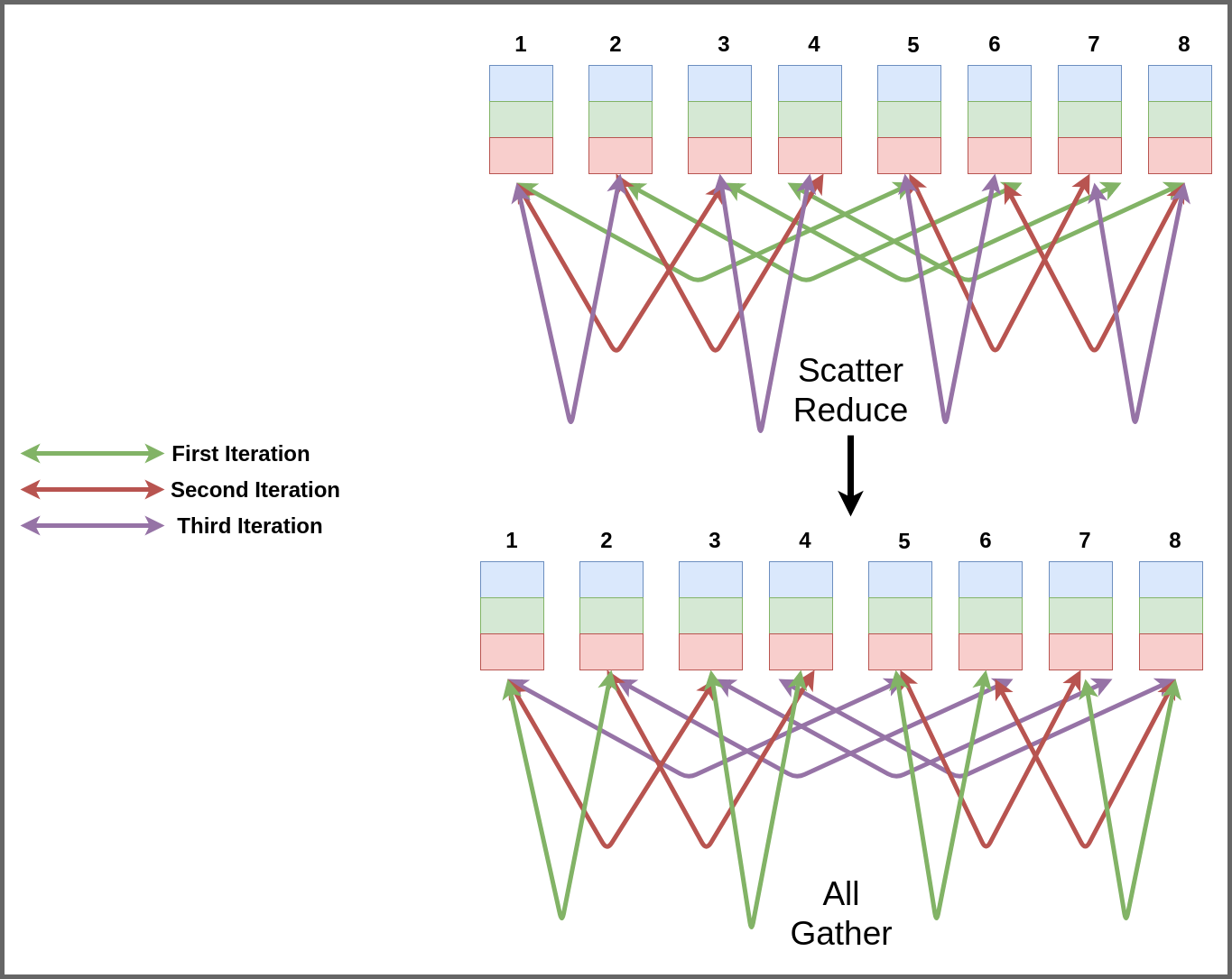}
\label{fig:model3}
\caption{Halfing and Doubling Algorithm}
\end{figure*} 

The recursive distance doubling and vector halfing algorithm in \cite{thakur2005optimization} works using 4 different primitives that are used in the algorithm. These are given as follows:

\begin{itemize}
  \item \textbf{Recursive Vector Halfing}: The vector is halfed at each time step.
  \item \textbf{Recursive Vector Doubling}: Small pieces of the vector scattered across processes are recursively gathered to form a large vector.
  \item \textbf{Recursive Distance Halfing}: The distance between machines is halfed with each communication iteration.
  \item \textbf{Recursive Distance Doubling}: The distance between machines is doubled with each communication iteration.
\end{itemize}
Similar to the ring algorithm, the All Reduce algorithm is made up of two procedures- the scatter-reduce and the all gather. The difference between this and the ring algorithm is in how these procedures perform the operation. The scatter reduce for recursive distance doubling and vector halfing algorithm runs for $log(P)$ steps, where $P$ is the number of processors and that $P$ is a power of two.

\subsection{Scatter Reduce}

\begin{algorithm}
\caption{Scatter Reduce  Vector Halfing Algorithm}
\begin{algorithmic}[1]
\Procedure{Scatter Reduce Vector Halfing}{}
\State $\textit{origVa} \gets \text{gradients computed by machine  }\textit{i}$
\State $rVal \gets \textit{orig}$
\State $dist \gets \textit{totMachs}/2$
\State $reVec \gets orig$ 
\For{ j = 0; j \textless distance; j++}:
\If{ ((i + dist) \% totMachs \textless i)}:
\State $sVec \gets topHalf(rVec)$ 
\Else
\State $sVec \gets botHalf(rVec)$ 
\EndIf
\State $sndMach(i + distance,sVec)$
\State $rVal \gets mach(i + dist)$
\State $rVec \gets rVec + rVal$
\If{ ((i + distance) \% totalMachines \textless i)}:
\State $rVec \gets botHalf(rVec)$
\Else
\State $rVec \gets topHalf(rVec)$ 
\EndIf
\State $dist \gets dist/2$
\EndFor
\EndProcedure
\end{algorithmic}
\end{algorithm}

\begin{itemize}
\item Machine $i$ communicates with machine $i + p/b$, where $b = 2$ in the first step and is multiplied by 2 after each step.
\item This communication between 2 machines happens as follows, machine $i$ divides it's vector into two parts. One part is used for sending and the other for receiving and reducing on. For example, if machine 1 could use the top half of the vector to send and the bottom part to receive and reduce on, then the second machine will use the opposite configuration.
\item After data is received from the counterpart process, the received data is used to reduce the original data. This reduced vector is halfed (vector halfing)  for the next step. Hence in the next step, distance between machines is $p/4$ and the data is halfed in the current step thereby doubling the distance and halfing the vector (recursive distance doubling and vector halfing).
\end{itemize}

If P is not a power of two, the algorithms are slightly modified by calculating the largest power of two less than $P$, this is denoted by $pz$. The value $r = P-pz$ is calculated and the first $2r$ machines are used to do the following:
\begin{itemize}

\item The even numbered machines communicate with the odd numbered machines i.e machine $i$ (where $i$ is even) communicates with machine $i+1$.
\item Both these machines, exchange data such that the even machines have the reduced vector of both.
\item Next, these even numbered machines are used along with the last $r$ machines in the recursive distance doubling and vector halfing algorithm described above, it is imperative to note that the odd numbered machines in the first $2r$ machines are not used in this procedure.
\end{itemize}

The above algorithm makes sure that the recursive distance doubling and vector halfing algorithm operates on a power of two number machines because the number of even numbered machines + $r$ is always a power of two

Once the scatter-reduce procedure is complete, each machine has a $1/p^{th}$ sized chunk of the final resultant vector. To broadcast these chunks on every machine, the all gather collective primitive is used which gathers data from each machine and broadcasts the resultant vector to each machine. The all gather for recursive distance doubling and vector halfing algorithm runs for $log(P)$ steps, where $P$ is the number of processors and communicates in the exact opposite way as the scatter reduce.

\subsection{All Gather}
\begin{itemize}

\item Machine $i$ communicates with machine $i + p/b$, where $b = 2^{log(P)}$ in the first step and is divided by $2$ after each step.

\item The communication between 2 machines happens as follows- machine $i$ divides its vector into two parts. The final chunk is meant for sending and rest of the data is replaced by the data that is received.

\item For the next step, the vector to be sent is the combination of the received chunk and the sent chunk, this is known as vector doubling as the vector doubles in size after each communication iteration.

\item This doubling of the vector size and halfing the distance between machines continues until $log(P)$ steps. One might notice that this is reverse of what happens in the scatter reduce process, the final result being with each machine having the final resultant vector. The time complexity for this algorithm is the same as the scatter-reduce.

\end{itemize}

\begin{algorithm}
\caption{All Gather  Vector Doubling Algorithm}
\begin{algorithmic}[1]
\Procedure{All Gather Vector Doubling}{}
\State $\textit{orig} \gets \text{gradients computed by machine  }\textit{i}$
\State $rVal \gets \textit{orig}$
\State $dist \gets 1$
\State $rVec \gets orig$ 
\For{ j = 0; j \textless totMach; j++}

\If{\textbf{if} ((i + dist) \% totMachs \textless i)}:
\State $sVec \gets topHalf(rVec)$
  \Else
\State $sVec \gets botHalf(rVec)$ 
 \EndIf
\State $sndMach(i + dist,sVec)$
\State $rVal \gets mac(i + dist)$
\State $rVec \gets concat(rVec, rVal)$
\If{((i + dist) \% totMachs \textless i)}
\State $rVec \gets replace(botHalf(rVec), rVal)$
  \Else
\State $rVec \gets replace(topHalf(rVec, rVal)$
 \EndIf
\State $dist \gets dist*2$
\EndFor
\EndProcedure
\end{algorithmic}
\end{algorithm}

After the all gather procedure all machines have the resultant vector signaling the end of the All Reduce process. The final complexity to the entire algorithm is $2*A*log{P} + 2*n*B$ where $A$ is the startup time per message, $B$ is the transfer time per byte and $n$ is the number of bytes transferred. That reduction procedure complexity is ignored as that is independent from the communication between machines. Hence, the final complexity for power of two number of processes is $2*A*logP + 2*n*B$ and for non power of two processes it is $2*A*logP + A + 3*n*B$. If the number of machines are not a power of two, the resultant vector needs to be sent to the odd numbered machines after the All Reduce ends which results in an overhead of $A + n*B$.

The advantages of using this algorithm is that the complexity of this operation is reduced from $2*A*P + 2*n*B$ to $2*A*logP + 2*n*B$, reducing the complexity of the algorithm from $\mathcal{O}(n)$ time to $\mathcal{O}(\log{}n)$ time. The disadvantages of using this algorithm is when the number of machines are not a power of two, a substantial overhead can be introduced as a number of machines (the first $2r$ odd numbered machines) are left unused during the All Reduce process, hence reducing the scalability of the program with respect to the total number of machines. The binary blocks algorithm which is described in the next section reduces this overhead.

\section{Binary Blocks Algorithm}
\label{sec:binary}

The binary blocks algorithm is an extension to the recursive distance doubling and vector halfing algorithm, it seeks to lower the degree of load imbalance for when the number of machines are not a power of two. In the original algorithm for the non power of two case, a number of machines are set aside until the algorithm completes its execution, after which they receive the resultant vector. This approach leads to a large number of machines being left idle in some cases, for example, for a cluster of 600 machines, 86 machines would be left idle while the processing executes on the remaining 512 machines. There is a significant load imbalance encountered in the network using this approach.

\begin{algorithm}
\caption{Binary Blocks Master Server Algorithm}
\begin{algorithmic}[1]
\Procedure{Binary Blocks Algorithm}{}
\State $\textit{blocks} \gets arrayList(blocks)$
\State $\textit{totalNumBlocks} \gets len(blocks)$
\BState \emph{\textbf{for} j = 0; j \textless totalNumBlocks; j++}:
\State $block \gets blocks[j]$ 
\State $scatterReduceBinaryBlocks(block,blocks)$
\BState \emph{\textbf{wait} scatterReduceAllEnd}:
\BState \emph{\textbf{for} j = 0; j \textless totalNumBlocks; j++}:
\State $block \gets blocks[j]$ 
\State $allGatherBinaryBlocks(block,blocks)$
\EndProcedure
\end{algorithmic}
\end{algorithm}

The binary blocks algorithm seeks to alleviate this problem by dividing the number of machines into blocks of power of twos. As an example, a 600 machine cluster will have 4 groups with $2^9, 2^6, 2^4, 2^3$ machines respectively. The steps of the binary blocks algorithm are outlined as such:
\begin{itemize}

 \item Each block executes the scatter-reduce procedure of the recursive distance doubling and vector halfing algorithm using the machines that are allocated to it. After a block finishes it's scatter reduce procedure, the machines in the smallest block send their reduced final chunk data to the machines of the block that is next in line in the size hierarchy. This data is reduced with the corresponding data on the machines of the bigger block. Here, the bigger block is signified by the block containing more number of machines.

 \item This data transfer and reduction after the scatter reduce is continued until the data reaches the biggest block. After the scatter reduce and transfer of data between all blocks has been completed, the reversal of the same process is started to distribute the final vector to all machines (the all gather procedure).

 \item Starting from the biggest block, data is sent down to the smaller blocks alongside the data transfer for the all gather procedure in their own block. Once a block gets data from a bigger block, it starts it's all gather procedure and transfers data to the block below. This process goes down the block hierarchy until the all gather completes on all blocks.

\end{itemize}

\begin{algorithm}
\caption{Scatter Reduce Master Client Algorithm}
\begin{algorithmic}[1]
\Procedure{Scatter Reduce Binary Blocks}{}
\State $\textit{curBlockNum} \gets i$
\State $\textit{curBlock} \gets block$
\State $\textit{allBlocks} \gets blocks$
\State $redVec \gets scatterReduce(curBlock)$ 
\BState \emph{\textbf{wait} scatterReduceEnd}: 
\BState \emph{\textbf{if} curBlockNum+1 \textless len(blocks)}:
\State $curBlockNum++$
\State $sendRedVecToBlock(curBlockNum, redVec)$ 
\EndProcedure
\end{algorithmic}
\end{algorithm}

The time complexity of the binary blocks algorithm is $2logP + 2nB$, the load balance depends on the amount of data transfer between machine inter block. This algorithm doesn't completely solve the load imbalance problem as there is a high transfer of data between  imbalanced blocks. However, it has been observed that the binary blocks algorithm works well even for $8+4$ and $16+8$ configurations in \cite{goyal2017accurate} making it a good alternate for clusters with non power of two number of machines.

\section{Fault Tolerant All Reduce}
\label{sec:faulttolerantreduce}

In a scenario where a distributed cluster consists of devices with low reliability, All Reduce algorithms need to restarted in case of a machine failure. We propose a fault tolerant version of the binary blocks algorithm by incorporating elements of the Raft consensus algorithm \cite{raft} into the All Reduce. This algorithm is resilient to machine failures and continues execution as long as backup replicas are operational.

   \begin{figure}[thpb]
      \centering
      \includegraphics[scale=0.4]{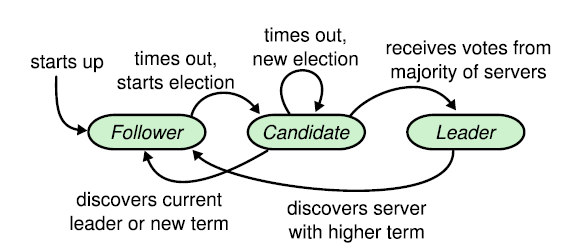}
      \caption{Raft Algorithm}
      \label{fig:raft}
   \end{figure}

\begin{itemize}
    \item Each participating node in Raft has a set of replicas attached to it. The number of replicas for each node usually 3 or 5 which is in line with what the authors of Raft recommend. These replicas maintain state among each other using Raft's replication algorithm. 
    \item Raft's replication algorithm elects a leader which accepts all state changes. In the case of the All Reduce, the reduction and concatenation operations required for the scatter reduce and all gather procedures are reflected as state changes. All the replicas listen in on the leader for state changes.
    \item If the leader goes down, the raft leader election protocol \cite{raft} is used to elect a new leader. Raft uses a combination of a randomized time out through which a replica applies for candidacy for leadership, the node that receives the majority of votes first is elected the new leader. If a candidate receives a request for a vote, it will immediately relinquish it's candidacy and vote for the prospective leader who requested the vote. The randomized timings allow for the lock step to be bypassed.
\end{itemize}

There have been several modifications proposed in the Raft protocol in systems like Zookeeper \cite{hunt2010zookeeper}, Kafka \cite{kreps2011kafka} to name a few. All of these consensus algorithms that maintain state come from the Paxos family of algorithms \cite{lamport2001paxos, chandra2007paxos}. Raft is popular for it's practicability and boasts the same fault tolerance and speed of a Multi Paxos system.
Since each node has some replicas, the replicas take the master's place in case of node failure, this results in a system which remains resilient to machine failures.

\begin{figure*}[htb]
\centering
\includegraphics[width=1.0\textwidth]{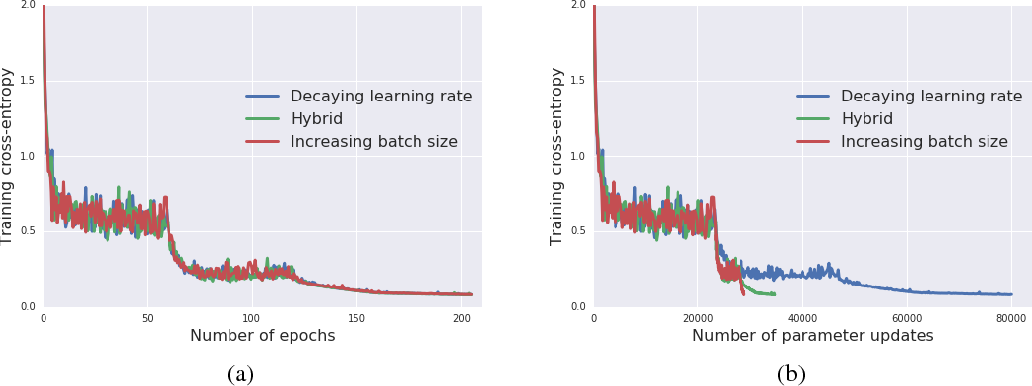}
\label{fig:model4}
\caption{Decaying Learning Rate vs Increasing Batch Size \cite{smith2017don}}
\end{figure*}

\section{Scaling Batch Size}

 In practice while training a deep neural network, the learning rate is slowly annealed as the training goes through various epochs. The intuition behind this is that the weights are allowed to take large steps at the beginning of training and smaller steps as the model is closer to convergence. This works quite well in practice and leads to a greater degree of convergence than a model which is trained with a fixed learning rate. It is also efficient as the large learning rate at the beginning can make a lot of progress before finetuning it with a smaller learning rate. However, training with large batch sizes is a promising avenue as it can speed up training time dramatically by lowering the training time from days to minutes as evidenced in \cite{goyal2017accurate, jia2018highly,  ginsburg2018large} and the work by \cite{smith2017don} reinforces this trend by empirically proving that increasing batch size is equivalent to decaying the learning rate

The benefit of training with bigger batch sizes is the lesser number of overall weight updates a model has to perform which leads to faster training as denoted by Fig 4. However training with large batch sizes naively, shows several issues with early divergence and lower final validation accuracy compared to a model trained on a smaller batch size \cite{goyal2017accurate, jia2018highly,  ginsburg2018large}. Training with a increasing batch size schedule works as follows:

\begin{itemize}
    \item Starting with a learning rate $l$ and batch size $b$, the batch size is increased by some factor after some epochs.
    \item The maximum batch size should be significantly smaller than the size of the training set. However, deciding on a good learning rate to use for the model can be non trivial and is dependant on the dataset.
\end{itemize}

A trick that can be used in deciding the learning rate is by using a learning rate finder \cite{lesliesuperconvergence}. A way to further decrease the number of parameter updates is to use the one cycle policy whilst using the increasing batch size scheme \cite{lesliesuperconvergence}. However, this could result in a loss of test accuracy, keeping the learning rate constant with increasing batch sizes is a safer way to train \cite{smith2017don}. Over the years, there has been a concerted effort to train neural networks with extremely large batch sizes. A large batch size has two benefits:

\begin{itemize}
    \item \textbf{Parallelization}: A larger batch size will enable more machines to train a model concurrently and help parallelize the process. This reduces the overall training time since machines work parallelly on training with a big batch size which reduces the total number of batches to train for an epoch to complete (Synchronous SGD).
    
    \item \textbf{True Distribution}: It has been observed that the gradients computed over a larger batch size match the true distribution of the final weights better than those with smaller batch sizes \cite{goyal2017accurate}.
\end{itemize}

However, there are several problems that are encountered while training networks with large batch sizes. Training seems to diverge after a certain threshold of batch size is passed, for example AlexNet diverges when a batch size of 2,000 is used \cite{goyal2017accurate, ginsburg2018large}. It has also been observed in \cite{goyal2017accurate, ginsburg2018large} that the final validation accuracy of the model begins to decrease as the batch size is increased indicating a decrease in the generalization capacity of the model. This phenomenon is also called the "generalization gap", several techniques have been proposed that seek to decrease this generalization gap and increase the batch size during training, these are looked at in the next sections.

\subsection{Linear Scaling Rule}
\label{sec:linearrule}

The linear scaling rule \cite{goyal2017accurate} is a simple technique that scales the learning rate with the batch size linearly.
 
\textbf{Linear Scaling}: \textit{Multiply the learning rate by $k$ when the mini-batch size is multiplied by $k$.}

\begin{itemize}

    \item \textbf{Scaling Rule}: Start with a learning rate $n$, and increase it gradually to $k*n$ where $k$ is the total number of machines over a period of 5 epochs. A small learning rate is used to warm up the training for a period of 5 epochs. As training gets distributed to more machines, the batch size is slowly raised to $k*n$ at the end of the $5^{th}$ epoch. This warm up step is crucial as gradients at the start of training are large and applying the linear scaling rule at the beginning leads to divergence.
\end{itemize}

The linear scaling rule allows for training of networks with batch sizes up to 8,192 \cite{goyal2017accurate}. It is important to note that this training doesn't suffer from a gap in validation accuracy compared to the single machine model. The work by \cite{goyal2017accurate} uses the recursive halfing and recursive doubling algorithm for the All Reduce showing that a $3\times$ speed improvement over the ring algorithm. A 90\% scaling efficiency was achieved via clever pipelining of the All Reduce operations (gradients being sent for aggregation as soon they're computed for a layer). The network is trained for 90 epochs irrespective of mini batch size and training finishes in an hour without any loss in accuracy with a cluster of 256 GPUs. It has been observed that the linear scaling rule doesn't allow for network to be trained for mini batches larger than 8,192 which The Layer Adaptive Learning Rates (LARS) \cite{ginsburg2018large} algorithm seeks to tackle via a novel concept. 

\subsection{LARS}
\label{sec:lars}

The linear scaling rule proposed by \cite{goyal2017accurate} allows for training the Resnet \cite{resnet} model with a batch size of 8,192. A large learning rate is proposed to accomodate for a smaller number of iterations due to a larger batch size. However in practice, training tends to diverge for large learning rates and a larger batch size results in a lower validation accuracy. As an example, the accuracy for AlexNet \cite{krizhevsky2012imagenet} for a batch size of 4,000 dips to 53.1\% from the baseline (B=256) of 57.6, increasing the batch size to 8,000 further dips the test accuracy to 44.8\% \cite{ginsburg2018large}. It is observed that applying batch normalization to AlexNet leads to a significant improvement in accuracy closing the gap to only 2.2\% from the previous 14\% for a batch size of 8,000.

The authors of \cite{ginsburg2018large} proposed using different learning rates for different layers of the neural network since it was observed that the ratio between the norm of the weights to the norm of the gradients is different for different layers. For example, in the AlexNet model, the ratio for the first conv layer is 5.76 while the ratio for the last fully connected layer is 1345. 

The local LR  $\lambda^l$ for each layer $l$ is computed as follows:

\begin{equation}
  \triangle w^l_t  =  \gamma* \lambda^l * \nabla L(w^l_t) 
 \end{equation}
 where $\gamma$ is a global LR.  Local LR  $\lambda^l$ is defined for each la    yer through "trust" coefficient $\eta < 1$:
\begin{equation}
   \lambda^l =  \eta  \times \frac{||w^l||}{||\nabla L(w^l)||}
    \label{eq:lars}
  \end{equation}
 The $\eta$ defines how much we trust the layer to change its weights during one update. Note that now the magnitude of the update for each layer doesn't depend on the magnitude of the gradient anymore, so it helps to partially eliminate vanishing and exploding gradient problems. This definition  can be easily extended for SGD to balance the local learning rate and the weight decay term $\beta$:

\begin{equation}
    \lambda^l = \eta  \times \frac{||w^l||} {||\nabla L(w^l)|| + \beta *||w^    l|| }
   \label{eq:lars_wd}
\end{equation}

The different learning rates are generated according to the ratio of the norm of weights and the norm of gradients for that layer. If that ratio is large, a high learning rate is computed and vice versa, this correlates with the observation that layers at the end of the network learn faster than those at the beginning. LARS seeks to modify learning rates throughout the training of the model according to the rate a layer is learning at that moment.

The usage of LARS allows training for batch sizes up to 64,000 with minimal loss in accuracy \cite{jia2018highly}. A small accuracy gap is observed, however that can be alleviated by training for a greater number of epochs. The accuracy gap is attributed to the fact that the stochastic gradients calculated over a large mini batch match the true gradients very closely. Presently, LARS is the state of the art in training with large batch sizes.

\section{Tensor Fusion}
\label{sec:tensorfusion}
 It has been observed for some popular models like the Resnet that the size of tensors computed for the gradients are quite small. More specifically, gradient tensor sizes for convolution layers
are much smaller than fully-connected layers. This information is particularly salient as sending small amounts of data over the wire can result in a substantial amount of latency overhead whilst simultaneously under-utilizing network bandwidth. A straight forward way of addressing this tensor fusion \cite{jia2018highly}, which is simply fusing multiple small tensors together to form a tensor of some minimum size before sending this fused tensor across the network. The benefits of performing this fusion is the reduction of the overhead of the startup time of each machine and overall reduction of the frequency if network traffic. This allows the networks to be clutter free and serve in optimal time. However, using tensor fusion for small tensors can lead to the ring All Reduce becoming inefficient and slow, \cite{jia2018highly} proposes a \textit{hierarchical All Reduce} that uses a multi layered master slave setup that is observed to give lower latencies. The hierarchical All Reduce works by splitting the total number of machines into batches after which each batch elects a master which aggregates the gradients. These masters perform the ring All Reduce among themselves, after which the masters distribute the updated gradients to their respective followers. This strategy side steps the overhead of the ring All Reduce overhead by reducing latency by a factor of the number of batches. Using tensor fusion reduces small network transactions and improves the overall speed of the network and is highly recommended. It's wide spread use in production systems like Horovord \cite{sergeev2018horovod} and Tencent's Framework \cite{jia2018highly} make it an important staple in modern distributed training frameworks.

\section{Low Precision Training}
\label{sec:training}

The fastest time it takes to train a Resnet model \cite{zhang2015deep} on the Imagenet database \cite{deng2009imagenet} as of this date is 4 minutes \cite{jia2018highly}. In that work, the authors use a combination of LARS algorithm, hybrid All Reduce algorithm, tensor fusion and mixed precision training on a cluster of 2048 GPU's connected with a low latency zero copy RDMA network connection. The hybrid All Reduce algorithm combines the ring All Reduce with the hierarchial version, switching between them depending on the tensor sizes at the current step.
A novel approach that they used  that enabled substantial gains in the overall speed of training was mixed precision training \cite{micikevicius2017mixed}. The increased throughput and bandwidth efficiency achieved due to it led to a speedup of training time by a factor of $8\times$.
As the name suggests, mixed precision training trains a neural network with two different data types- a smaller data type for a majority of operations and a bigger data type for precision critical operations. Neural networks have originally used single or double precision numbers as their default data type as these data types worked well in capturing the representational capacity of the task the network wanted to model. Single precision numbers are 32 bit floats and double precision numbers are 64 bit floats. Recent research suggests that the speed and size of a neural network can be reduced by upto 50-80\% by training on lower precision data types \cite{krishnamoorthi2018quantizing,jacob2017quantization}. A popular approach is to train a network using 16 bit floats training (FP16 training) however these networks have reported an inferior test accuracy than their single precision counterparts \cite{micikevicius2017mixed}. This happens primarily due to the weight updation step that happens in low precision. More specifically, multiplying the low precision gradients with the learning rate can sometimes result in the number overflowing the 16 bit range leading to an incorrect calculation that in turn leads to a loss of final validation accuracy.

Mixed precision training \cite{micikevicius2017mixed} seeks to solve this by using a single precision (32 bit) master copy of weights and running everything else in half precision (16 bit). The process works as follows:
\begin{itemize}
    \item The master copy of the weights is kept in the single precision format. These weights are converted to the half precision format after which the forward and backward pass is run on these 16 bit weights to compute the gradients

\item When the gradients have been received, they are converted to a single precision format (32 bits) and the weight updation step is performed using these single precision weights where the gradients are multiplied by the learning rate and added to the old weights to form the new set of weights for the master copy.
    \item These new weights are stored as the new master copy and this process continues on for the next batch of data.

\end{itemize}

Mixed precision training has enabled higher throughput hence enabling the reduction of the computation vs communication bottleneck. There are however some caveats to mixed precision training that one needs to be aware of, namely loss drop off and inferior arithmetic precision

\subsection{Loss Scaling}
Using mixed precision training yields to divergence in training for some network architectures. This is because the low precision operand exponent bias centers the range of normalized value exponents to [-14, 15] while gradient values in practice tend to be dominated by small magnitudes (negative exponents). It is observed that a big portion of the 16 bit subspace is left unused by the gradients, many of the gradients are below the minimum representable range and become zeros \cite{micikevicius2017mixed}. 

This can be fixed by scaling the floating point values such that they occupy the entire representable range. For example, for a SS  class network \cite{redmon2017yolo9000}, multiplying gradients by a factor of 8 allows for training and resultant accuracy equal to that of the original single precision version. There are a variety of approaches that can be used in deciding the scaling factor. One efficient way is to scale the loss value computed in the forward pass, doing this scales the gradients too via the backpropogation step. There is one slight change to the weight updation step when loss scaling is used, that is weight gradients
need to be unscaled before the weight update to maintain the update magnitudes similar to that of single precision training. It is recommended to perform the weight unscaling right after backpropogation and before any other gradient related operations such as gradient clipping and weight decay among others are computed

Choosing a scaling factor can be performed using a multitude of options. One popular option is to choose a constant scaling rate, although a caveat of this being that there is a notion of hit and trial in choosing a scaling rate as one has to make sure that the value of the maximum gradient value multiplied by the scaling factor doesn't exceed the value of the maximum value of a low precision (16 bit) operand. To handle this case when a gradient overflow is detected for a batch during training, that batch is simply skipped and training moves on to the next batch.

\subsection{Arithmetic Precision}

It is observed that the result of the operation of two low precision (16 bit) operands with each other needs to be stored as a single precision (32 bit) operand before converting to a low precision operand and flushing to memory. Failure to do so results in a lower degree of convergence.  It is recommended that large reductions (sums across elements of a vector) should be carried out in single precision. Such reductions mostly come up in batch-normalization layers when accumulating from the statistics and softmax layers.
Both these layer types read and write low precision tensors from and to memory, however they perform the arithmetic in single precision. This did not slow down the training process since these layers are memory-bandwidth limited and are not sensitive to arithmetic latency. Since arithmetic precision doesn't impact speed of these operations, either low precision or single precision math can be used. 

 In conclusion, mixed precision training is an important technique for speeding up neural networks and specifically pertains to distributed learning where optimization in both the computation and the communication medium is needed.

\section{Gradient and Parameter Compression}
\label{sec:compression}

One of the primary bottlenecks in scaling distributed training process is the high bandwidth cost for communicating model weights and gradients between nodes. This bottleneck is even more significant when training on devices, especially for mobile devices using federated learning \cite{federated} which suffer from low network bandwidth and a slow connection. To this end, researchers have proposed various approaches to utilize network bandwidth efficiently. Using the asynchronous and synchronous variants of SGD allows nodes to communicate independently while parallelizing and improving network bandwidth utilization to an extent (\cite{dean2012large, recht2011hogwild, li2014communication}) but there have been significant advances in  gradient compression that show promising results \cite{lin2017deep}. These approaches are primarily based on two ideas:

1. \textbf{Quantization}: Gradient Quantization focuses on compressing gradients into efficient data representations by reducing the number of bits required per parameter and thus the overall gradient size to be communicated across the network. Recent work has shown great reduction in bandwidth cost without losing significant accuracy. Work from \cite{1bitsgd} proposed 1-bit SGD with error feedback which quantizes gradients into its sign (1-bit) while accumulating errors locally for the next batch, it shows upto $10 \times$ speed-ups with a small accuracy loss for Speech DNNs. Alistarh et al. \cite{qsgd} proposed Quantized SGD (QSGD) which quantizes components to discrete set of values without losing statistical properties using stochastical rounding to generate lossless encoding of the output. Using a 4-bit QSGD, the authors trained a 62M AlexNet on 8 GPUs resulting in speed-ups of $2.05 \times$ with a minor increase in Top-1 accuracy, similar gains were seen while training 13M LSTM \cite{greff2017lstm} on the AN4 dataset. Another technique TernGrad \cite{wen2017terngrad}, also uses stochastical rounding by quantizing gradients to ternary levels \{-1, 0, 1\} and proposes a layer-wise ternarizing and gradient clipping procedure to shrink gradient bounds. Experiments using TernGrad show that networks such as AlexNet with larger communication-to-computation ratios, tend to benefit more with speed-ups of upto $3.04 \times$ while retaining accuracies. However for GoogleNet, where the ratio is small, the speedup achieved is comparatively less with a 2\% decrease in accuracy \cite{wen2017terngrad}.

2. \textbf{Sparsification}: Not all parameters of neural networks change at once during training,  Gradients computed are found to be sparse, hence only a small number of weights need to be updated after each batch. Bandwidth cost can significantly be reduced if we leverage this observation and limit communication of all gradients across the network. Initial work based on this idea includes the work from Strom \cite{Strom2015ScalableDD} which proposed a static thresholding of gradients i.e communicating gradients once they are larger than a constant value. This thresholding resulted in speed gains of $54 \times$ with a 1.8\% reduction in error on an automatic speech recognition task. A compression ratio of $846-2871 \times$ was achieved when model size was increased from 14.6M to 48.8M parameters. However, this static threshold was a hyperparameter and suffered from tuning problems. Gradient Dropping by Aji et al. \cite{aji2017sparse} focuses on setting the threshold using a drop ratio (hyperparameter) by dropping \textit{R}\% of small gradients. This ratio can be selected for each node (local drop ratio) or set as a global drop ratio with layer normalization. Experiments resulted in a 49\% speedup on MNIST and a 22\% speedup on Neural Machine Translation by exchanging $50 \times$ less data without accuracy loss. Dryden et al. \cite{dryden2016communication} work put forward Adaptive Quantization where a fixed proportion of positive and negative updates are sent after processing a mini-batch. Results were close to that of 1-Bit quantization in terms of compression and accuracy but were fastest compared to other methods ($1.76 \times$ faster than next implementation which used no compression). However, these results were shown for only fully-connected models using the MNIST dataset. AdaComp \cite{chen2017adacomp} exploits local gradient activity by dividing the residual vector of every layer into several fixed size bins using the maximum absolute value in each bin as a threshold to find salient gradients. If the previous residue combined with the scaled calculated gradient exceeds this maximum value, gradients are considered important and are hence quantized before sending. Experimental results show close to a $40 \times$ compression rate for convolutional layers and a $200 \times$ compression rate for fully-connected and recurrent layers without a loss in accuracy and convergence.

\begin{algorithm}
\caption{Deep Gradient Compression for vanilla momentum SGD on node k}
\renewcommand{\algorithmicrequire}{\textbf{Input:}}
\begin{algorithmic}[1]
\Require $\text{dataset \textit{X}}$
\Require $\text{minibatch size \textit{b} per node}$
\Require $\text{momentum \textit{m}}$
\Require $\text{the number of nodes \textit{N}}$
\Require $\text{optimization function \textit{SGD}}$
\Require $\text{initial parameters \textit{w} = \{\textit{w}[0], \dots , \textit{w}[\textit{M}]\}}$
\State $\textit{U\textsuperscript{k}} \gets 0, \textit{V\textsuperscript{k}} \gets 0$
\For{\textit{t} = 0, 1, \dots}
   \State $G^k_{t} \gets 0$
   \For{\textit{i} = 1, \dots, \textit{b}}
       \State $\text{Sample data \textit{x} from \textit{X}}$
       \State $G^k_{t} \gets G^k_{t} + \frac{1}{\textit{Nb}}\triangledown f(x,w\textsubscript{t})$
   \EndFor
   \If{\text{Gradient Clipping}}
   \State $G^k_{t} \gets Local\_Gradient\_Clipping(G^k_{t})$
   \EndIf
   \State $U^k_{t} \gets m \cdot U^k_{t-1} + G^k_{t}$
   \State $V^k_{t} \gets V^k_{t-1} + U^k_{t}$
   \For{\textit{j} = 0, \dots, \textit{M}}
   \State $thr \gets s\% \text{ of } |V^k_{t}[j]|$
   \State $Mask \gets |V^k_{t}[j]| > thr$
   \State $\tilde{G}^k_{t} \gets V^k_{t}[j] \odot Mask$
   \State $V^k_{t}[j] \gets V^k_{t}[j] \odot \neg Mask$
   \State $U^k_{t}[j] \gets U^k_{t}[j] \odot \neg Mask$
   \EndFor
   \State $\textbf{All\-reduce: } G_{t} \gets \sum_{k=1}^{N}encode(\tilde{G}^k_{t})$
   \State $w\textsubscript{t+1} \gets SGD(w\textsubscript{t}, G\textsubscript{t})$
\EndFor
\end{algorithmic}
\end{algorithm}

Recently, Lin et al. proposed Deep Gradient Compression (DGC) \cite{lin2017deep} which takes cues from previous work in order to reduce the communication bandwidth greatly while preserving accuracy.

\begin{itemize}
    \item Similar to previous approaches, DGC sends gradients based on a threshold after encoding and accumulating the rest of the gradients locally for the next iteration. 
    \item Since sparse updates harm convergence, techniques such as momentum correction and local gradient clipping \cite{lin2017deep} are applied  to retain accuracy.
    \item SGD with Nestrov Momentum performs better for convergence compared to vanilla SGD, however it cannot be applied here directly since it ignores the discounting factor introduced by thresholding. To fix this, instead of accumulating gradients locally, the authors propose accumulating velocity, $U_{t}$. This is known as momentum correction. Also, to avoid the exploding gradient problem, gradient clipping is used.
    \item Gradient thresholding delays small gradients updates which can lead to a stale gradient problem which harms model convergence. Applying a threshold mask to the accumulated gradients and momentum factor limits the momentum of delayed gradients, thereby reducing staleness. A learning rate warm-up is used while training to allow the network to smoothen the rapid changes in early stages of training. Also, gradient sparsity is exponentially increased from a small value to a final value to allow training to adapt to the high sparsity in the gradients.
\end{itemize}
DGC was tested on image classification, language modeling and speech recognition tasks and compared with previous work. It achieved a compression ratio of $597 \times$ on AlexNet and a $277 \times$ on the ResNet-50 with a slight increase in accuracy and no loss in convergence. For language modeling, DGC achieves a compression ratio of $462 \times$ and a $608 \times$ on speech recoginition tasks with a slight reduction in perplexity and word error rate respectively. It is currently the state of the art in gradient compression.

\section{Future Work}
\label{sec:future}

The field of distributed training has seen great progress in the past few years and is ripe for further innovation. The fields of interest currently are increasing the limit of the mini batch size to which a network can train on without diverging or reducing final validation accuracy for Synchronous SGD and researching ways to solve the stale gradients and lower final validation accuracy problems for Asynchronous SGD. 
Training on lower powered devices has gained some momentum in the form of federated learning \cite{federated} with several modern deep learning frameworks building blocks for secure and decentralized training.
Training on consumer devices has several advantages, one of them being able to build intelligent applications that learn specialized habits based on customer interaction while keeping customer data private through on device storage and training. An example of such an application is the Android Predictive Keyboard which learns a small personalized language model for next word predictions by training on device.  A key challenge to training on low powered devices is the low network bandwidth and compute available. An efficient framework can unlock mass training on phones and IOT devices thereby unlocking portable applications to deep learning. Some promising work has been done by \cite{chahal_mathur_2018} for distributing training on low powered devices through a reinforcement learning algorithm to schedule training jobs on a heterogeneous cluster of devices. Unlocking distributed training on commodity devices needs various optimizations in training and communication. Gradient compression and mixed precision training are a few directions that have shown good results and hold fruitful promise. Overall, the field seems to have an active and innovative research direction and is primed to be a core component for enabling widespread intelligent applications. 

\section*{Conclusion}

We have surveyed and summarized the various components of a distributed training framework and recommend the following techniques to build an efficient and scalable distributed training framework. 

\begin{itemize}
    \item  Using the Synchronous SGD algorithm is recommended for training due to its strict convergence guarantees.
    
    \item  The binary blocks algorithm should be used for the All Reduce procedure for gradient accumulation due to its superior running time.
    
    \item To use hardware and network bandwidth efficiently, various techniques such as gradient compression, quantization and mixed precision training should be utilized in the framework. We recommend using a combination of Deep Gradient Compression and Mixed Precision Training.
    
    \item Training should be performed using extremely large batch sizes to maximize parallelizability and minimise running time. We recommend the LARS algorithm as it has proven to be robust enough to train networks with batch sizes upto 64,000.  
    
\end{itemize} We also propose a fault tolerant All Reduce algorithm that works without complete restarts in an unreliable environment. Finally, we mention some of the directions and implications that future work in distributed training could take.

\bibliographystyle{IEEEtran} 
\bibliography{bibliography.tex} 

\end{document}